\documentclass[letterpaper, 10 pt, conference]{ieeeconf}  

\IEEEoverridecommandlockouts                              

\usepackage{graphicx}
\usepackage{amsmath}
\usepackage{amsfonts}
\usepackage{nicefrac}
\usepackage{comment}
\usepackage{hyperref}
\usepackage{xspace}
\usepackage{multirow}
\usepackage{subcaption}
\usepackage{tabularx}
\usepackage{booktabs}
\usepackage[table,xcdraw]{xcolor}

\usepackage{verbatimbox}

\let\labelindent\relax
\usepackage{enumitem}

\usepackage{blindtext}
\usepackage{graphbox}

\usepackage[ruled,vlined,linesnumbered]{algorithm2e}

\usepackage{amssymb}
\usepackage{pifont}

\DeclareMathOperator*{\argmax}{arg\,max}

\renewcommand{\baselinestretch}{0.99}

\makeatletter
\DeclareRobustCommand\onedot{\futurelet\@let@token\@onedot}
\def\@onedot{\ifx\@let@token.\else.\null\fi\xspace}
 
\def\ie{\emph{i.e}\onedot}

\makeatother

\title{\LARGE \bf
Domain Randomization for Robust, Affordable and Effective\\ Closed-loop Control of Soft Robots
}

\author{Gabriele Tiboni$^{1,\dagger}$, Andrea Protopapa$^{1,\dagger}$, Tatiana Tommasi$^{1}$, and Giuseppe Averta$^{1}$
\thanks{*This work was supported by Politecnico di Torino, CINECA, HPC@POLITO. This study was carried out within the project FAIR - Future Artificial Intelligence Research - and received funding from the European Union Next-GenerationEU (PIANO NAZIONALE DI RIPRESA E RESILIENZA (PNRR) – MISSIONE 4 COMPONENTE 2, INVESTIMENTO 1.3 – D.D. 1555 11/10/2022, PE00000013). This manuscript reflects only the authors’ views and opinions, neither the European Union nor the European Commission can be considered responsible for them. $\dagger$Gabriele Tiboni and Andrea Protopapa contributed equally to this work (corresponding
author: Gabriele Tiboni, e-mail: {\tt\small gabriele.tiboni@polito.it})}
\thanks{$^{1}$Politecnico di Torino, Turin, Italy {\tt\small first.last@polito.it}}%
}

\begin{document}

\maketitle
\thispagestyle{empty}
\pagestyle{empty}

\begin{abstract}
Soft robots are gaining popularity thanks to their intrinsic safety to contacts and adaptability. However, the potentially infinite number of Degrees of Freedom makes their modeling a daunting task, and in many cases only an approximated description is available. 
This challenge makes reinforcement learning (RL) based approaches inefficient when deployed on a realistic scenario, due to the large domain gap between models and the real platform. 
In this work, we demonstrate, for the first time, how Domain Randomization (DR) can solve this problem by enhancing RL policies for soft robots with: i) robustness w.r.t. unknown dynamics parameters;
ii) reduced training times by exploiting drastically simpler dynamic models for learning;
iii) better environment exploration, which can lead to exploitation of environmental constraints for optimal performance. 
Moreover, we introduce a novel algorithmic extension to previous adaptive domain randomization methods for the automatic inference of dynamics parameters for deformable objects.
We provide an extensive evaluation in simulation on four different tasks and two soft robot designs, opening interesting perspectives for future research on Reinforcement Learning for closed-loop soft robot control.
\end{abstract}

\section{INTRODUCTION}
\label{sec:introduction}
Soft robotics is a rapidly developing field that has the potential to revolutionize how robots interact with their environment~\cite{kim2013soft}. Unlike their rigid counterparts, soft robots are made from materials that can deform and adapt to their surroundings, enabling them to perform novel and unprecedented tasks in fields such as healthcare~\cite{runciman2019soft} and exploration \cite{aracri2021soft}.
\begin{figure}[t]
    \centering
    \includegraphics[width=\linewidth]{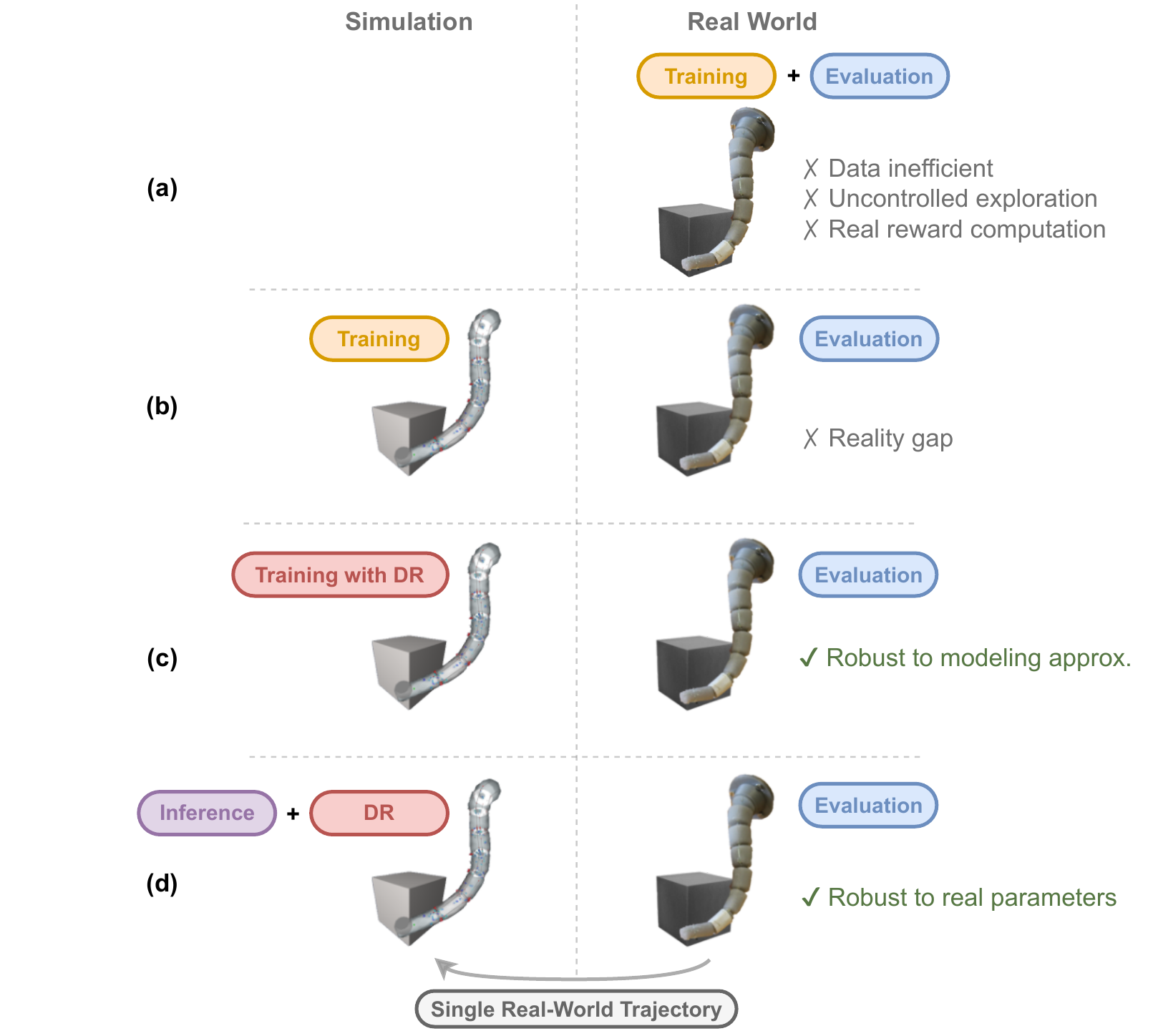}
    \caption{
    Paradigms in RL-based robot learning: a) training directly on the real world; b) na\"ive Sim-to-Real transfer suffers from the reality gap; c) Training with domain randomization increases robustness to modelling approximations and errors; d) distributions over simulator dynamics parameters may be automatically inferred from real-world data for use with DR.}
    \label{fig:paper_intro_overview}
    \vspace{-0.5cm}
\end{figure}
However, controlling continuous soft robots is a challenging task: an accurate kinematic model would require infinite degrees of freedom (DoF)~\cite{dubied_sim--real_2022}, and their highly nonlinear dynamics are difficult to model realistically~\cite{renda2014dynamic}. In addition, novel actuation mechanisms have introduced further challenges to approach optimal closed-loop control of soft robots. Popular designs include (1) cable-driven soft robots, which employ cables that can be extended or retracted to control the robot's shape and motion, or (2) pneumatic-based models, which rely on the pressurization of air chambers within the soft robot.

Many attempts have been made to control soft devices through model-based techniques, also pushed by the advancement of modelling techniques \cite{della2021model}. Yet, most complex tasks appear to be still unfeasible without the use of data-driven learning methods \cite{kim2021review}. 
In particular, Reinforcement Learning has been showing promising results in recent years to learn effective closed-loop control policies for soft robots \cite{bhagat_deep_2019,thuruthel2019model}, avoiding the need to know the exact dynamics of the system.
Due to the notoriously low sample efficiency of current RL algorithms, these methods often rely on learned forward dynamics models~\cite{thuruthel2019model,centurelli2022closed}, or on simulated models to train a policy that can later be transferred to the real world for evaluation~\cite{schaff2022soft, li_towards_2022} (See Fig.~\ref{fig:paper_intro_overview}.b). The latter approach has been gaining more traction thanks to recent advances in accurate and efficient simulators for deformable objects~\cite{coevoet2017software, naughton2021elastica}. This line of research is commonly denoted as \textit{Sim-to-Real Transfer}, which already demonstrated remarkable results for rigid robots in recent years~\cite{peng2018sim,tan2018sim}.
However, the performance of learned policies in the real-world often falls short of expectations due to the differences between the simulation and the real-world, namely policies are affected by the \textit{reality gap}. This is further exacerbated by the complexity of measuring deformable object parameters and designing accurate modeling, limiting current applications in soft robotics to overly simplified models for locomotion environments~\cite{schegg2022sofagym,schaff2022soft} or simple trajectory tracking tasks~\cite{li_towards_2022}.
Therefore, we identify the need to design novel sim-to-real transfer methods to scale the training of soft robots to more complex tasks, such as manipulation and contact-rich settings, and to overcome modeling approximations.

To bridge the reality gap, Domain Randomization (DR) has been proposed as a promising technique to learn transferable policies for rigid robotic systems~\cite{peng2018sim,tan2018sim,zhao2020sim_sim2realsurvey}. DR involves training policies on simulated environments with randomized dynamics parameters sampled from a predefined distribution, promoting a policy that is robust to variations (see Fig. \ref{fig:paper_intro_overview}-c). Recent studies have also attempted to automatically estimate DR distributions~\cite{tsai_droid_2021,ramos2019bayessim,tiboni2022dropo}, a method referred to as Adaptive Domain Randomization (ADR) (see Fig. \ref{fig:paper_intro_overview}-d).
While DR and ADR have been successful in improving the transferability of policies for rigid robots, their effectiveness in soft robotics remains largely unexplored.

In this work, we present a thorough investigation of Domain Randomization in the context of closed-loop soft robot control.
We include the examination of existing DR techniques for learning robust control policies on recently proposed soft robot benchmark environments~\cite{schegg2022sofagym, goury_fast_2018}---namely a reaching task for a trunk robot and walking for a MultiGait soft design. Moreover, we design a novel extension to state-of-the-art ADR methods and evaluate its capabilities to infer complex deformable object parameters.
Finally, we propose two novel challenging manipulation setups in simulation for the cable-driven trunk robot (\emph{pushing} and \emph{lifting}) which we release to the public.

Our findings demonstrate that our method may accurately infer complex dynamics parameters such as Poisson's Ratios and Friction coefficients, and lead to policies that are robust to parameter discrepancies among domains. Furthermore, we discover that such policies can even be learned using simpler modeling approximations in simulation, yielding affordable and drastically reduced training time complexity. Finally, we test the capabilities of Domain Randomization to improve the task efficiency altogether, by acting as a regularization effect on the exploration of the environment. Interestingly, we notice that randomizing the surrounding environment allows the agent to find more effective strategies to solve the same task, exploiting task-specific environmental constraints.

\section{RELATED WORK}
\label{sec:related_work}
\subsection{Soft Robotics: modeling, actuation and simulation} 
As previously anticipated, describing continuous soft robots is a challenging task, because their modeling lays in the domain of continuum mechanics. Soft robots are actuated devices usually composed by viscoelastic material (such as silicone). Their dynamics, therefore, is regulated by infinite-dimensional Partial Differential Equations. Interestingly, recent works have demonstrated that finite-dimensional
approximations of the robot’s dynamics provide a reasonable trade-off between model tractability and accuracy \cite{della2021model}. The most popular designs implement either pneumatic or cable-driven actuation mechanisms \cite{schegg2022review}. The first consist of inflatable chambers which, when filled with air or fluids, may change their length, curvature and shape. The latter, instead, present cables or other extensive elements attached to specific points of the robot body. Cables are then actuated through external bodies, producing pushing or pulling forces on the insertion points. A proper combination of multiple actuation elements (i.e. different chambers or cables actuation schemes) can provide complex open loop robot behaviors. 

Based on different modeling strategies (reviewed e.g. in \cite{della2021model, schegg2022review} and here omitted for the sake of space), several engines have been proposed so far to provide tools for an efficient and effective simulation of soft robotic devices. Among the others, it is worth mentioning SOFA~\cite{coevoet2017software}, ChainQueen~\cite{hu2019chainqueen}, Abaqus~\cite{narang2018mechanically}, which use volumetric FEM techniques, and Elastica~\cite{naughton2021elastica}, SimSOFT~\cite{grazioso_geometrically_2018}, and SoRoSim~\cite{Mathew2022SoRoSim} that, instead, leverage on discretization of rod models.

\subsection{Sim-to-Real Transfer with Domain Randomization}
Domain randomization has become a popular approach for transferring learned policies from simulation to real-world hardware for \textit{rigid} robotic systems~\cite{zhao2020sim_sim2realsurvey}. Such approach has been widely investigated both for randomizing the visual appearance of the simulator~\cite{tobin2017domain,james2017transferring,sadeghi2016cad2rl}, and its dynamics parameters~\cite{peng2018sim,tan2018sim}. In these cases, the goal is to learn a policy that is invariant to changes in state space or transition dynamics, respectively.
However, training a single policy to perform well on overly large variations of the environment may not always be possible, opening the challenge to design sensible posterior distributions over dynamics parameters~\cite{muratore2022robot,vuong2019pick}. In \cite{peng2018sim}, the authors proposed to use memory-based policies to allow the agent to readily adapt its behavior through implicit dynamics inference, while increasing complexity at training time. Alternatively, Adaptive Domain Randomization (ADR) methods attempt to automatically estimate DR distributions over dynamics parameters of interest, e.g. object masses and friction coefficients.
Methods such as DROID~\cite{tsai_droid_2021}, DROPO~\cite{tiboni2022dropo}, and BayesSim~\cite{ramos2019bayessim} fall in the latter category, and have been recently shown to produce effective inference and promising policy transfer results for rigid robotic systems. These ADR methods move from the iterative-based online counterparts~\cite{chebotar_closing_2019}, as they don't interact with real hardware at optimization time and can make use of off-policy collected data---see~\cite{tiboniadrbenchmark} for reference on online vs. offline ADR approaches.
Critically, DROID is confined to rigid robotics due to its reliability on joint torques measurements for position-controlled systems. Similarly, DROPO makes strict assumptions that prevent it from being applied to partially-observable environments---i.e. configurations of deformable bodies. 

Despite recent advances, Domain Randomization has yet to be thoroughly investigated in the context of soft robotic systems.
Two initial attempts are conducted by injecting random noise in the state observations~\cite{centurelli2022closed}, and more recently to both observations and policy actions~\cite{li_towards_2022}. However, although encouraged by the community~\cite{schaff2022soft,schegg2022sofagym}, no randomization of dynamics parameters for parametric physics engines currently exists. The same statement applies to the investigation of ADR methods for parameters inference of deformable bodies.
We aim to fill this gap by sheding light on the novel challenges concerning the application of DR on infinite-DoF systems, providing evidence for its effectiveness and designing a novel algorithmic modification to DROPO to cope with partially-observable environments.

\section{BACKGROUND}
\label{sec:background}
\subsection{Reinforcement Learning}
Consider a discrete-time dynamical system described by a Markov Decision Process (MDP) $\mathcal{M}$, with state space $\mathcal{S}$, action space $\mathcal{A}$, initial state distribution $\mu(s_0)$, transition dynamics probability distribution $\mathcal{P} (s_{t+1}|s_t, a_t)$ and reward function $r(s_t, a_t)$. At each time $t$, the \textit{environment} $\mathcal{M}$ evolves according to the current state $s_t \in \mathcal{S}$ and action $a_t \in \mathcal{A}$ taken by an \textit{agent}, i.e. the decision maker, with initial state drawn according to $\mu(s_0)$. Denote as $\pi_\theta(a_t|s_t)$ the stochastic \textit{policy} used by the agent to interact with the environment, parameterized by $\theta$. Under this formulation, Reinforcement Learning (RL) addresses the problem of finding an optimal policy $\pi_\theta^*(a|s)$ such as to maximize the expected (discounted) cumulative reward:
\begin{equation}
\label{eq:rl_goal}
\pi_{\theta}^{*} =\underset{\pi_\theta }{\argmax } \  \mathbb{E}_{\pi_{\theta}, \mathcal{P}, \mu} \left[ \ \sum _{t=0}^{T} \gamma ^{t} r(s_t, a_t)\right]
\end{equation}
with discount factor $\gamma \in (0, 1]$. For complex problems with a continuous state space $\mathcal S$, the policy $\pi_\theta$ is in practice parameterized by a neural network $\theta$, learned, e.g., through policy gradient RL algorithms such as Proximal Policy Optimization (PPO)~\cite{schulman2017proximal}.

\subsection{Learning from randomized simulators}
\label{subsec:dr_background}
In the sim-to-real transfer paradigm a simulator is used as training environment, referred to as the \textit{source domain} and denoted with $\mathcal{M}_s$. We assume $\mathcal M_s$ to share the same state space and action space of the real-world environment $\mathcal M_r$, noted as the \textit{target domain}. In contrast, the source environment is further parameterized by its dynamics parameters $\xi \in \mathbb R^{n_\xi}$, such as masses, friction coefficients, and Poisson's ratio of deformable objects, which ultimately affect the source transition dynamics $\mathcal{P}_\xi$.
Dynamics parameters can be generally assumed to be random variables that obey a parametric distribution $p_\phi(\xi)$, parameterized by $\phi$. In a domain randomization setting, the agent's goal is therefore to learn a policy that maximizes~(\ref{eq:rl_goal}) while acting in a randomized environment according to $p_\phi(\xi)$: 
\begin{equation}
\label{eq:rl_goal_withdr}
\pi_{\theta}^{*} =\underset{\pi_\theta }{\argmax } \ \mathbb{E}_{\xi \sim p_{\phi } (\xi )}\left[ \mathbb{E}_{\pi_\theta, \mathcal{P}_{\xi } ,\mu}\left[ \ \sum _{t=0}^{T} \gamma ^{t} r(s_t, a_t)\right]\right]
\end{equation}

In practice, DR may be easily integrated into existing RL algorithms by randomly sampling new dynamics parameters $\xi \sim p_\phi(\xi)$ at the beginning of each training episode.

\subsection{Adaptive Domain Randomization}
We consider here the \textit{offline} Adaptive Domain Randomization paradigm, i.e. the more general case of ADR where no assumptions are made on how target domain data is collected for the inference phase---hence suitable to work with off-policy data or human demonstrations. In particular, let $\mathcal{D}$ be a dataset of state-action transitions $\mathcal{D}=\{(s_0, a_0, s_1), \dots , (s_T, a_T, s_{T+1}) \}$, previously collected on the target domain. Under this formulation, ADR methods introduce an \textit{inference phase} prior to policy optimization, that aims to find the optimal $p^*_{\phi}(\xi)$ domain randomization distribution given $\mathcal{D}$. Later, $p^*_{\phi}(\xi)$ is used to train a policy using DR with the objective in~(\ref{eq:rl_goal_withdr}), which can ultimately be transferred to the target domain for evaluation.

\section{RESET-FREE DROPO}
\label{sec:method}
We describe our novel algorithmic modification to the current state-of-the-art offline ADR method DROPO~\cite{tiboni2022dropo}, to cope with partial observability in soft robotic environments.

\subsection{Method overview}
By design, Domain Randomization Off-Policy Optimization (DROPO) relies on replaying real-world data in simulation, by resetting the simulator to each visited real-world state and executing the same corresponding action. On one hand, this allows the method to avoid the rising of compounding errors when replaying real trajectories~\cite{tiboniadrbenchmark}. On the other hand, this approach assumes that target observations encode full information on the configuration of the scene, such that the internal state of the simulator may be reset to each state. While this may be a reasonable assumption in rigid robotics, where robot state is typically observable at all times, it prevents the algorithm from being applied to partially-observable environments. In other words, a soft robot manipulator may only be tracked by a discrete number of points, but its exact infinite-DoF configuration is too complex to encode, hence impossible to recover in simulation.
In this work, we relax this algorithmic assumption and develop an extension named \textit{Reset-Free} (RF) \textit{DROPO}.  Our novel implementation draws inspiration from DROID~\cite{tsai_droid_2021}, where only the initial full configuration of the environment is assumed to be known, and actions are replayed in open-loop during inference. In contrast to DROID, however, we retain the likelihood-based objective of DROPO which does not suffer from converging to point-estimates---as in~\cite{tsai_droid_2021}---due to inherent convergence properties of the evolutionary search algorithm CMA-ES~\cite{hansen2006cma}---see claims in~\cite{tiboni2022dropo,tiboniadrbenchmark}. Finally, we introduce a novel regularization technique to allow our method to effectively deal with compounding errors and obtain informative likelihood estimates. Following the original pipeline, RF-DROPO ultimately consists of (1) a data collection phase where the dataset $\mathcal{D}$ is made available, (2) an inference phase where $p_\phi(\xi)$ is optimized to maximize the likelihood of real-world data to occur in simulation, and (3) policy training with the converged domain randomization distribution.

\begin{algorithm}[t]
\SetAlgoLined  
 \SetKwInOut{Input}{input}
\SetKwInOut{Output}{output}
 \Input{Initialize $\phi$, sequence \textit{NonDecrSeq, $\tau_{0}=1$}}
 \Output{Parameters $\phi^*$ of $p_{\phi^*}(\xi)$}
 Collect a dataset of transitions $D=\{(s_{t},a_t, s_{t+1})\}_{t=0}^{T}$ from the target domain\; \label{as:collect}
 \For{$i=0, \dots, M$ opt. iterations}
 { \label{as:opt_iterations} 
  Sample a set $\Lambda$ of $\min(\tau_i, L)$ transitions from $\mathcal{D}$; 
  \ForAll{$(s_{t}, a_t, s_{t+1}) \in \Lambda$}
  { \label{s:for_state}
    Estimate $S_{t+1}^{\phi} \sim \mathcal{P}_\phi(\cdot|s_0, a_{0,\dots,t}; \phi)$ through repeated sampling; \\ 
    Compute likelihood estimate of $s_{t+1}$ under $S_{t+1}^{\phi}$: $\mathcal{L}_t =  \mathcal{P}_\phi ( S_{t+1}^{\phi}=s_{t+1} |s_{0} ,a_{0,\dotsc ,t} ;\phi)$
  }
  $\mathcal{L} = \sum_t \log\mathcal{L}_t$; \\
  Update $\phi$ towards maximizing $\mathcal{L}$;  \label{as:update_phi} \\
  $\tau_{i+1} \leftarrow \min(T, \ \textit{NonDecrSeq}(\tau_{i}))$;
 }
 Train a policy with DR using the converged $p^{*}_{\phi} (\xi)$ ; \label{as:train_policy} \\
 \caption{Reset-Free DROPO}
 \label{alg:rfdropo}
 \vspace{-2pt}
\end{algorithm}

\subsection{Implementation}

\subsubsection{Data collection}
A fixed dataset $\mathcal{D}$ with target domain state-action transition shall be made available to run the inference phase. Such data may be collected with any desirable strategy, such as kinesthetic teaching or hard coded policies.
Note that no reward computation in the real-world is needed, as the inference phase relies on state-action transition pairs only.

\begin{figure}[t]
    \centering
    \includegraphics[width=0.8\columnwidth]{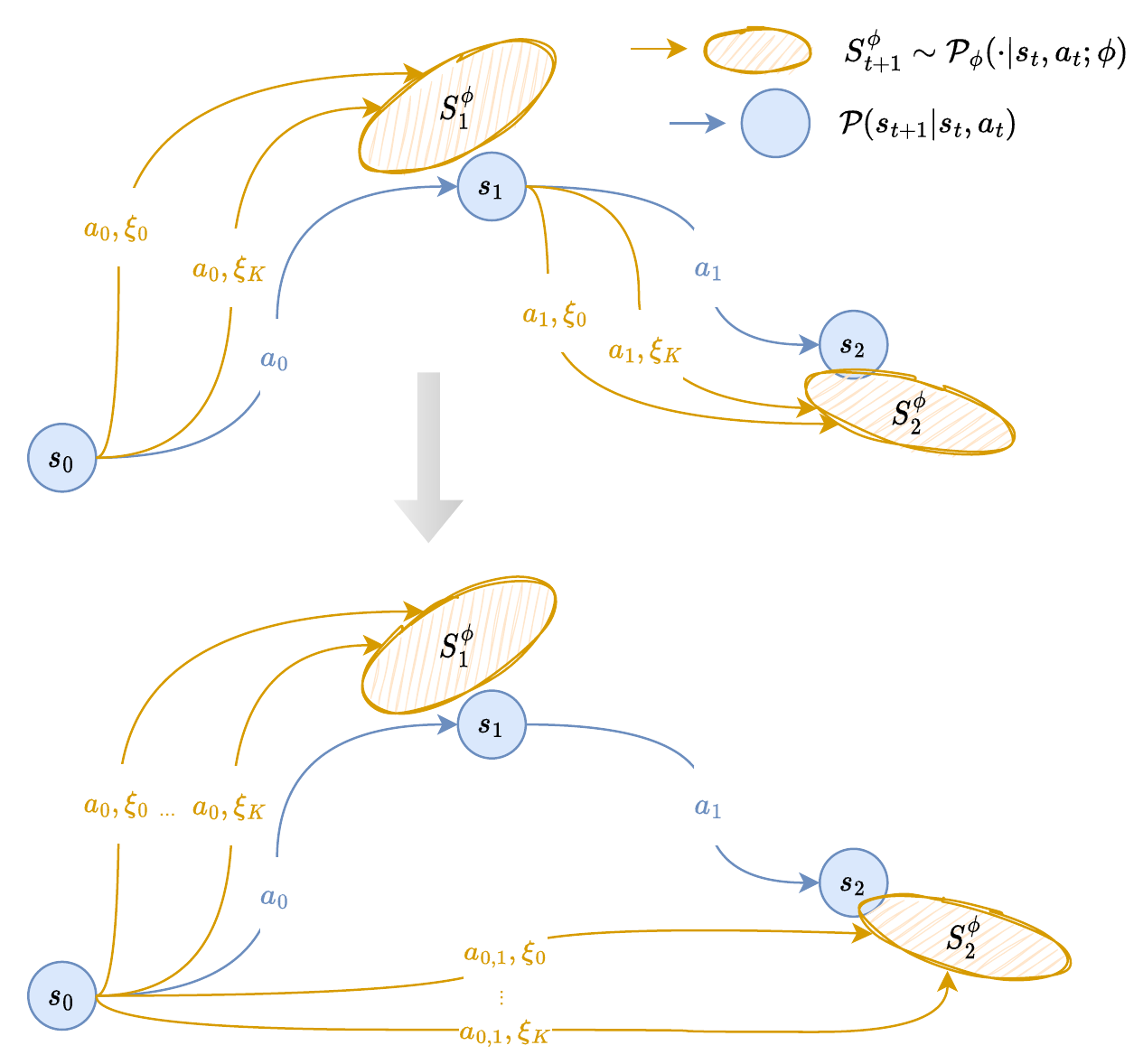}
    \caption{Overview of (bottom) Reset-Free DROPO algorithm vs. (top) its original counterpart with intermediate state-resetting.
    }
    \label{fig:dropo_vs_rfdropo}
    \vspace{-6pt}
\end{figure}

\subsubsection{Dynamics parameter inference}
Our objective follows the original DROPO implementation, as we seek to maximize the likelihood of real-world state transitions $s_{0,\dots,T}$ according to the simulation transition dynamics $\mathcal{P}_\xi$, under randomized parameters $\xi \sim p_\phi(\xi)$.
Therefore, let $S_{t+1}^{\phi}$ be the random state variable distributed according to the new transition dynamics $\mathcal{P}_\phi(\cdot | s_t, a_t; \phi)$, with stochasticity induced by the domain randomization distribution $p_\phi(\xi)$.
Our modification implies the avoidance of resetting the simulator state to each $s_t \in \mathcal{D}$. Instead, the probability of observing a target state $s_t$ is estimated through the execution of all preceding actions $a_{0, \dots, t-1}$, starting from the initial known configuration $s_0$---assumed to be the same as in the real-world.
Note how, for $t=1$, RF-DROPO and DROPO coincide, whereas longer-horizon likelihood computations differ. An illustration of the proposed algorithmic modification is depicted in Fig.~\ref{fig:dropo_vs_rfdropo}. Overall, RF-DROPO maximizes the following objective function, acting upon the DR distribution $\phi$:
\begin{equation}
\label{eq:rf_dropo_objective}
    \phi^{*} =\underset{\phi }{\argmax } \ \sum _{t=0}^{T} \log\mathcal{P}_\phi ( S_{t+1}^{\phi}=s_{t+1} |s_{0} ,a_{0,\dotsc ,t} ;\phi )
\end{equation}
where the \textit{log-}likelihood is considered for better numerical stability. In practice, the likelihood function---which models the relationship between dynamics parameters and state transition dynamics---is assumed to be unknown, as physics simulators act as black-box non-differentiable systems. The quantity in~(\ref{eq:rf_dropo_objective}) is then estimated independently for each time step by sample estimates, i.e. repeatedly observing and inferring the sim dynamics distribution $\mathcal{P}_\phi$ for different values $\xi_k \sim p_\phi(\xi)$, as in the original implementation.
It is worth noting that the likelihood computation for different timesteps $t$ is still independent, assuming i.i.d. sampling of $\xi_k \sim p_\phi(\xi)$ for different time steps and noting that a function---simulator dynamics--- of i.i.d. random variables still leads to statistically independent random variables.
For the sake of simplicity, we stick to uncorrelated multivariate Gaussian distributions as parameterization for $p_\phi(\xi)$, with additive homoschedastic variance $\epsilon$ to all dimensions---See Sec.~3.4 in~\cite{tiboni2022dropo}.

In the attempt to stabilize likelihood estimation for long-horizon time steps---which could suffer from compounding errors when executing multiple consecutive actions---we introduce a regularization temperature parameter $0\leq\tau \leq T$. We then propose to limit the time-horizon of our objective function~(\ref{eq:rf_dropo_objective}) to $\tau$---instead of $T$---and set $\tau$ to gradually increase during each optimization iteration until finally reaching the full trajectory length $T$. Such addition allows the algorithm to focus on a smaller number of state transitions in the near-future at the beginning of the inference phase, guiding the process to optimize dynamics parameters that capture gradually further state-transitions. In principle, any non-decreasing sequence of $\tau$ may be adopted; throughout our experiments, we use an exponential schedule $\tau ( i) =T\left( 1-e^{-i\ \cdot \ ( 2\log 10/M)}\right)$ for each opt. iteration $i=0,\dots,M$, which reaches 90\% of the total number of transitions $T$ half-way through the process.

Finally, without loss of generality, we allow the algorithm to consider only a random sub-sample of $L$ transitions at each CMA-ES iteration, among the total number of $\tau$ transitions considered within the current time horizon. Note how such extension draws inspiration from Stochastic Gradient Descent (GD) vs. vanilla GD, allowing to reduce inference time.
The overall modifications of RF-DROPO to the original implementation are summarized in Algorithm~\ref{alg:rfdropo}.

\begin{figure}[t]
    \centering
    \includegraphics[width=\linewidth]{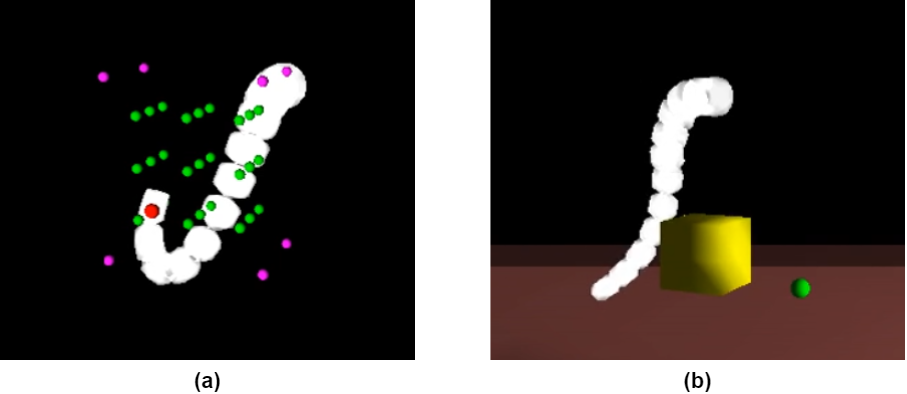}
    \caption{TrunkReach and TrunkPush setups. a) Purple dots define the box of possible goal locations sampled at training time, green dots are 27 fixed target locations for evaluation, with the red dot being the current goal. b) Green dot is the desired target location for the box center of mass.}
    \label{fig:trunk_reach_push}
    \vspace{-12pt}
\end{figure}

\subsubsection{Policy Training}
The converged distribution $p_{\phi}^*(\xi)$ obtained through the inference phase may be finally used to train a policy with Domain Randomization. We train our policies through Proximal Policy Optimization (PPO)~\cite{schulman2017proximal} across all our experiments.

\section{EXPERIMENTS}
\label{sec:experiments}
We carry out a thorough experimental evaluation in simulation on the effectiveness of Domain Randomization in the context of soft robotics. In particular, we aim at answering the following research questions:
\begin{itemize}[leftmargin=*]
    \item can RF-DROPO infer accurate posterior distributions over dynamics parameters for deformable objects?
    \item is Domain Randomization capable of efficiently transferring policies learned from simplified dynamics models, hence reducing training time and modeling complexity? 
    \item can Domain Randomization lead to more effective strategies through better exploration, e.g. by exploiting environmental constraints?
\end{itemize}

\begin{table}[b]
\caption{Posterior parameter estimation for the TrunkReach task. Search space is reported in $[min, max]$ format. Numerical values RF-DROPO and BayesSim (MDNN) are reported in mean and standard deviation $(\mu \pm\sigma)$.}
\centering
\resizebox{\columnwidth}{!}{%
\begin{tabular}{lclll}
\toprule
\textbf{} 
   &
  \multicolumn{1}{c}{\textbf{\begin{tabular}[c]{@{}c@{}}Trunk Mass \\ {[}$kg${]}\end{tabular}}} &
  \multicolumn{1}{c}{\textbf{Poisson's Ratio}} &
  \multicolumn{1}{c}{\textbf{\begin{tabular}[c]{@{}c@{}}Young’s Modulus\\ {[}$kg/(mm \cdot s²)${]}\end{tabular}}} \\ \midrule
\textbf{Target}                          & \multicolumn{1}{c}{0.42} & \multicolumn{1}{c}{0.45} & \multicolumn{1}{c}{4500} \\ \midrule
\textbf{Search space} &
  
  \multicolumn{1}{c}{{[}0.005, 1{]}} &
  \multicolumn{1}{c}{{[}0.4, 0.5{]}} &
  \multicolumn{1}{c}{{[}2000, 7000{]}} \\ \midrule
\textbf{BayesSim }  & 0.32 $\pm$ 1.28E-01           & 0.49 $\pm$ 2.30E-03           & 1914.25 $\pm$ 7.31E+06        \\ 
\textbf{RF-DROPO}       & 0.64 $\pm$ 1.00E-05           & 0.45 $\pm$ 1.00E-05           & 6878.01 $\pm$ 2.77E-02        \\ 
\bottomrule
\end{tabular}%
}

\label{tab:trunk_reach_inf}
\end{table}

\subsection{Tasks}

We test our method by building on recently introduced benchmark tasks in the domain of soft robotics for manipulation and locomotion~\cite{schegg2022sofagym}. In particular, we consider four evaluation domains for the purpose of our analysis: \textit{TrunkReach}, \textit{TrunkPush}, \textit{TrunkLift}, \textit{MultiGait}. The Trunk robot~\cite{goury_fast_2018} is a cable-driven continuum deformable manipulator, depicted in Fig.~\ref{fig:trunk_reach_push}. Its configuration is encoded in a 63-dimensional vector of positional keypoints along the robot, used as input observation vector for RL agents. The action space consists of a discrete set of $16$ actions, encoding the extension or retraction of one of the $8$ cables. 

\begin{figure}[tb]
    \centering
    \includegraphics[width=\linewidth]{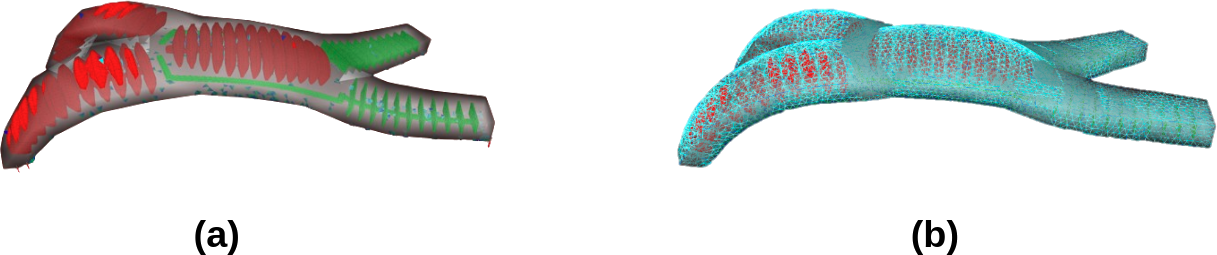}
    \caption{Multigait locomotion environment: (a) simplified vs. (b) accurate simulation model.}
    \label{fig:multigait}
    \vspace{-0.4cm}
\end{figure}

First, we test the trunk's capabilities in a reaching task, where the agent's goal is to reach a randomly located point in space with the robot's endpoint (\textit{TrunkReach}). Then, we design two novel manipulation settings to analyse the effects of DR in contact-rich settings: pushing a cube to a desired target location (\textit{TrunkPush}), and lifting a flat object in the presence of a nearby wall (\textit{TrunkLift})---see Fig.~\ref{fig:trunkwall_seq_1}. 
Finally, we consider the \textit{multi-gait} design~\cite{goury_fast_2018}, a complex pneumatic soft robot tasked with walking forward (see Fig.~\ref{fig:multigait}).
The latter setting introduces critical challenges by means of modeling approximations, which can drastically reduce the training time at the cost of higher modeling errors. In this context, we investigate the ability of DR to overcome modeling approximations and allow more affordable models to be used at training time. To do this, we instantiate two multi-gait models, the first (see \ref{fig:multigait}-b) is the original detailed model, while the second (see \ref{fig:multigait}-a) is a simplified version obtained through Model Order Reduction \cite{goury_fast_2018}.

All DR-compatible benchmark tasks, together with an implementation of our method, are publicly available at \url{https://andreaprotopapa.github.io/dr-soro/}.

\subsection{DR for robustness: parameter discrepancies}
\begin{figure}
\centering
\begin{subfigure}{0.49\columnwidth}
  \centering
  \includegraphics[width=1\linewidth]{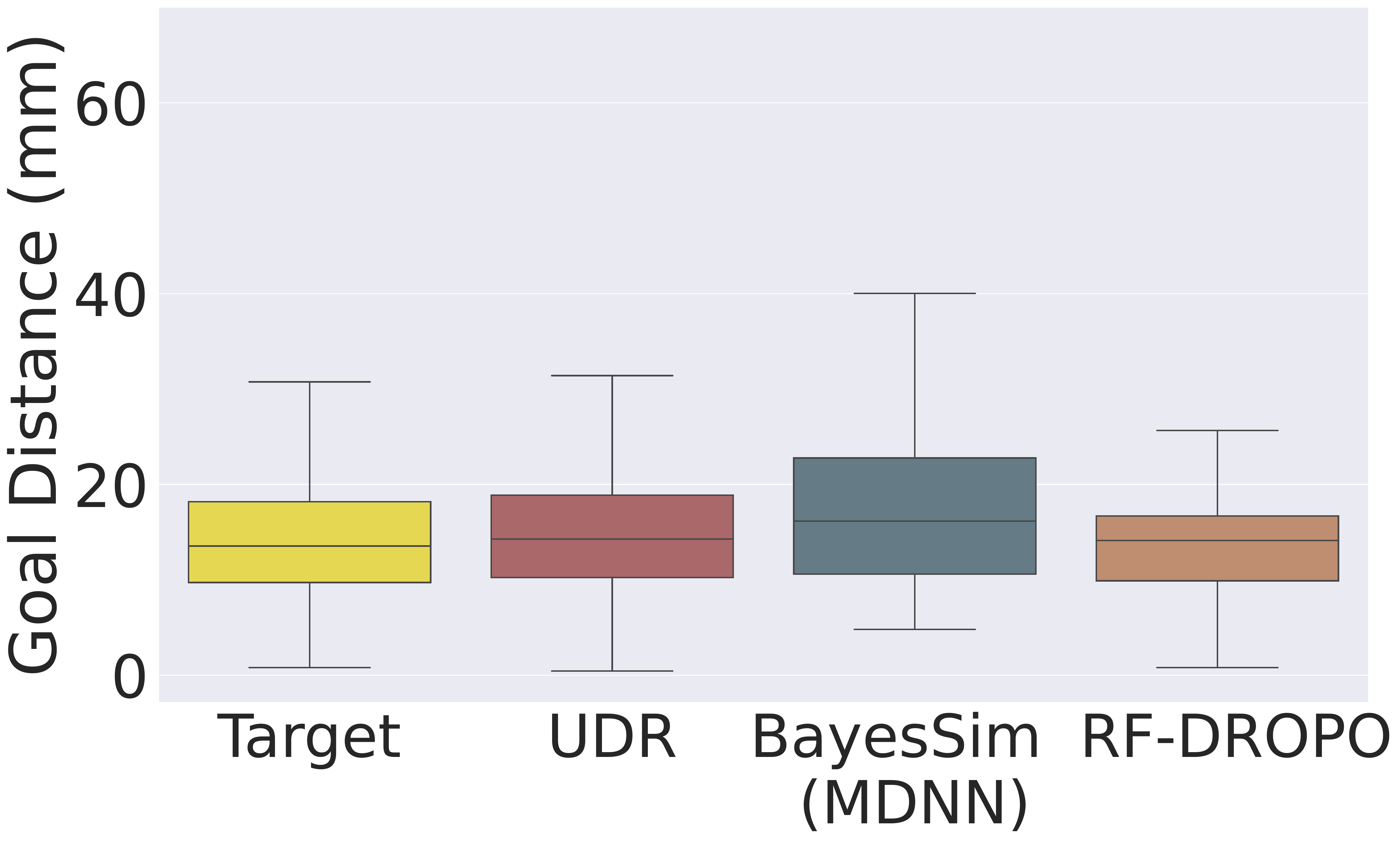}
  \caption{TrunkReach}
  \label{fig:sub1}
\end{subfigure}%
\begin{subfigure}{0.49\columnwidth}
  \centering
  \includegraphics[width=1\linewidth]{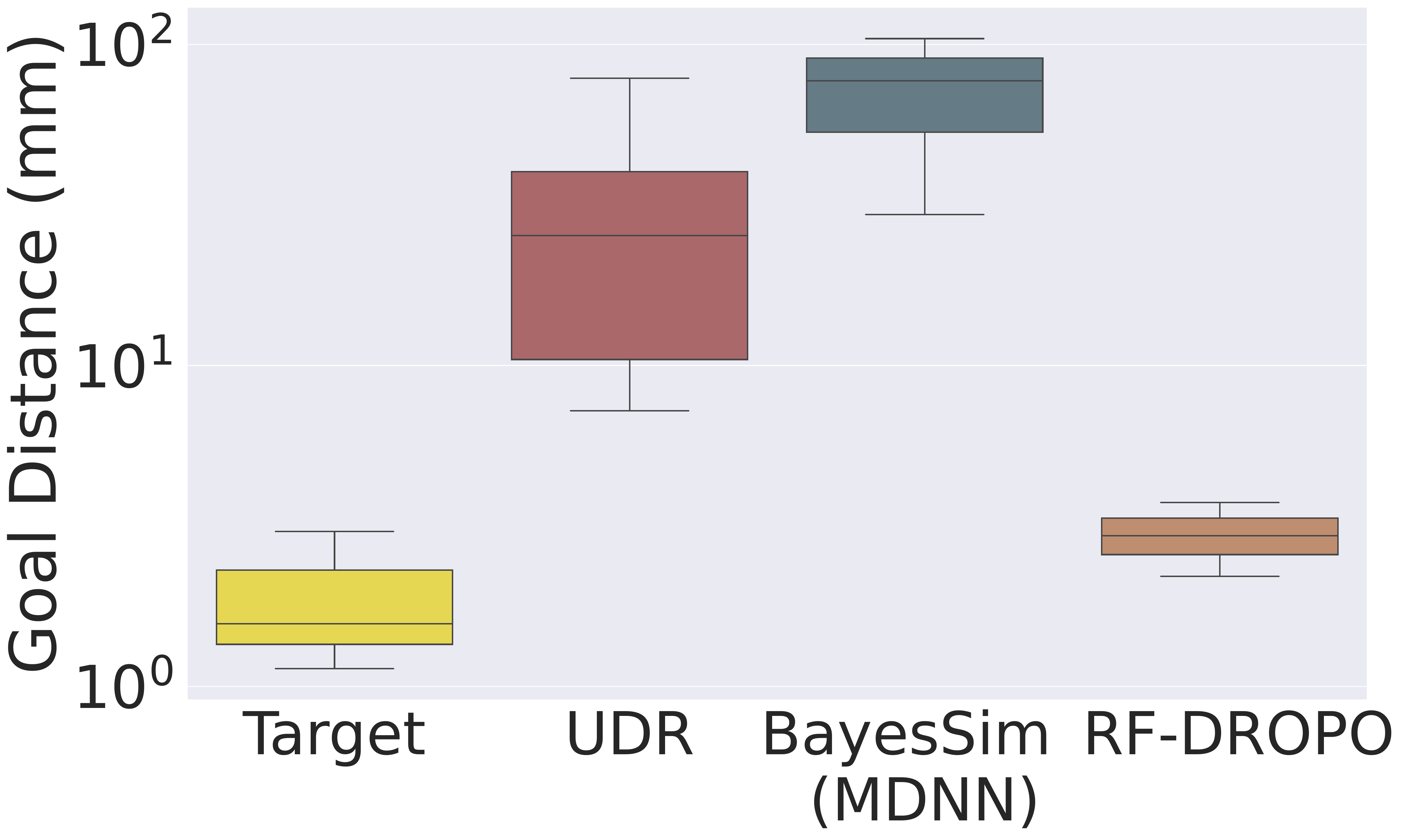}
  \caption{TrunkPush}
  \label{fig:sub2}
\end{subfigure}
\vspace{-2pt}
\caption{Vanilla parameter estimation: policy evaluation in terms of distance from the goal position (lower is better).}
\label{fig:trunkcube_trunkpush_resultsvanilla}
\vspace{-8pt}
\end{figure}

\begin{figure}[tb]
    \centering
    \includegraphics[width=0.5\linewidth]{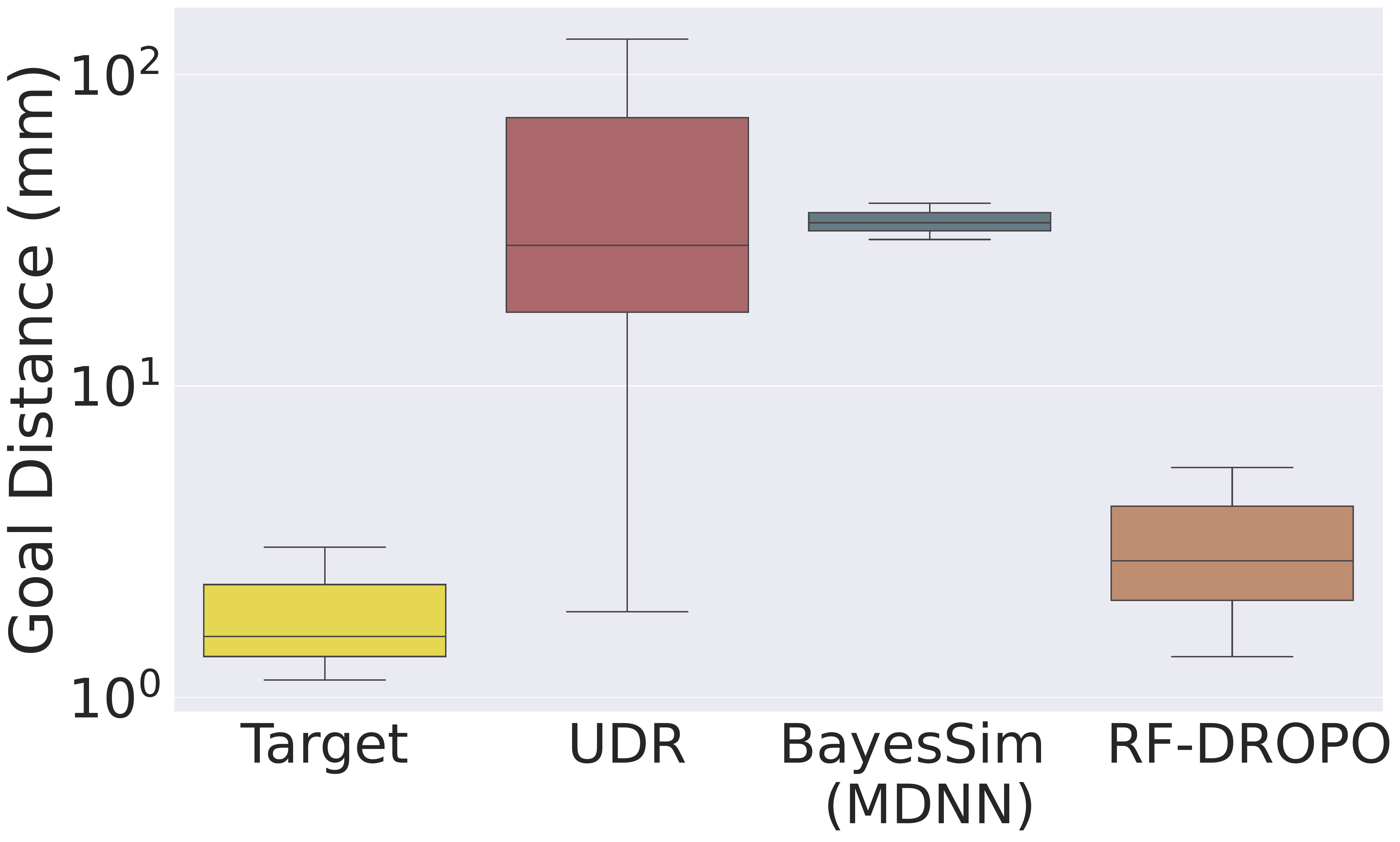}
    \vspace{-2pt}
    \caption{Unmodeled parameter estimation setting for the TrunkPush task. Evaluation on the nominal target domain in terms of distance from the goal position (lower is better).}
    \label{fig:trunkcube_unmodeleled}
    \vspace{-12pt}
\end{figure}

We compare the capabilities of RF-DROPO and the more established method BayesSim~\cite{ramos2019bayessim} to infer complex deformable object parameters in simulation. We consider the TrunkReach and TrunkPush tasks, and feed the respective methods with a single trajectory---the equivalent of 5 seconds in wall time---collected on each environment, by rolling out a semi-converged policy. In principle, collecting more data may increase the accuracy of the inference phase, but we found one trajectory to be sufficient while limiting the computational time for running RF-DROPO to roughly the same as policy optimization ($\approx$48 hours on 20 CPU cores).
It's also worth noting that target data in $\mathcal{D}$ does not need to contain high-reward state-transition pairs, but simply informative data to estimate the desired parameters.
We present our inference results in Tab.~\ref{tab:trunk_reach_inf} and~\ref{tab:trunk_push_inf}, respectively for the TrunkReach and TrunkPush tasks.
In particular, we experiment with both MDNN and MDRFF features of BayesSim and report the best results for brevity---as uncorrelated approximation of the full Gaussian distributions. Noteworthy, although a Gaussian mixture model is returned by BayesSim, a single modality was often found to be dominant.
In general, MDRFF provided less stable results with occasional unfeasible parameter values. 
Interestingly, RF-DROPO demonstrated accurate and reliable estimation of the Poisson's Ratio and Friction coefficient to impressive precision, whereas BayesSim fell shorter.
On the other hand, accurate inference of the Young's Modulus and respective masses proves to be more challenging in general, likely due to parameter correlations.
Later, we investigate whether policies trained with Domain Randomization on the estimated posterior distributions may transfer well to the target domain with nominal parameter values. For each method and task, three policies are trained with three different repetitions of the inference phase, and finally evaluated (see Fig.~\ref{fig:trunkcube_trunkpush_resultsvanilla}).
In addition, we compare the results with a Uniform Domain Randomization (UDR) baseline, which reflects the performance of 10 policies trained on uniform DR distributions randomly sampled from the search space in Tab.~\ref{tab:trunk_reach_inf} and~\ref{tab:trunk_push_inf}.
Notably, although the estimated parameters do not perfectly match, the performance of RF-DROPO on the target domain are still comparable to that of an oracle policy directly trained on it. On the other hand, BayesSim fails to transfer effectively in the TrunkPush task, likely due to a poorer posterior estimation---which is pivotal for contact-rich tasks.
In particular, note how a random search over sensible uniform distributions (\ie UDR) may produce well-performing policies for a simple reaching task, yet fails on average when contacts and more complex dynamics are involved.
Finally, we design a more challenging \textit{unmodelled setting} for the TrunkPush task where the Young's Modulus is misspecified by 80\% and excluded from the estimation process of ADR methods. In this context, DR distributions should then be inferred to compensate for such misidentified parameter. The final performance of compared ADR methods is reported in Fig.~\ref{fig:trunkcube_unmodeleled}, with their respective estimated posteriors in Tab.~\ref{tab:trunk_push_inf}. Similarly to the vanilla setting, we notice the performance gap between RF-DROPO and BayesSim, with the former consistently pushing the box about $8mm$ away from the target location on average over three repetitions.

\renewcommand{\arraystretch}{1.2}
\begin{table*}[]
\caption{Parameter estimation in the vanilla and unmodelled settings for the TrunkPush task.}
\vspace{-4pt}
\begin{tabular}{llcccccc}
\hline
 &
  \textbf{} &
  \multicolumn{1}{l}{\textbf{}} &
  \textbf{\begin{tabular}[c]{@{}c@{}}Cube Mass \\ {[}$kg${]}\end{tabular}} &
  \textbf{\begin{tabular}[c]{@{}c@{}}Friction \\ Coefficient\end{tabular}} &
  \textbf{\begin{tabular}[c]{@{}c@{}}Trunk Mass \\ {[}$kg${]}\end{tabular}} &
  \multicolumn{1}{l}{\textbf{Poisson's Ratio}} &
  \textbf{\begin{tabular}[c]{@{}c@{}}Young's Modulus\\ {[}$kg/(mm \cdot s^2)${]}\end{tabular}} \\ \cline{2-8} 
 &
  \textbf{Target} &
  \textbf{} &
  0.05 &
  {\color[HTML]{24292F} 0.3} &
  0.42 &
  0.45 &
  4500 \\ \cline{2-8} 
 &
  \textbf{Search space} &
  \multicolumn{1}{l}{{[}min, max{]}} &
  {[}0.005, 1{]} &
  {[}0.01, 1{]} &
  {[}0.005, 1{]} &
  {[}0.4, 0.5{]} &
  {[}2000, 7000{]} \\ \hline
 &
  \textbf{BayesSim} &
  $(\mu \pm \sigma)$ &
  0.37 $\pm$ 9.43E-02 &
  0.37 $\pm$ 1.07E-01 &
  0.69 $\pm$ 1.29E-01 &
  0.42 $\pm$ 3.50E-03 &
  1855.10 $\pm$ 8.17E+06 \\
\multirow{-2}{*}{\textbf{Vanilla}} &
  \textbf{RF-DROPO} &
  $(\mu \pm \sigma)$ &
  0.06 $\pm$ 4.40E-03 &
  0.30 $\pm$ 1.51E-03 &
  0.52 $\pm$ 3.24E-03 &
  0.45 $\pm$ 4.50E-04 &
  5557.07 $\pm$ 2.45E+00 \\ \hline
 &
  \textbf{BayesSim} &
  $(\mu \pm \sigma)$ &
  0.52 $\pm$ 8.96E-02 &
  0.51 $\pm$ 8.29E-02 &
  0.48 $\pm$ 7.47E-02 &
  0.44 $\pm$ 9.00E-04 &
  {\color[HTML]{CB0000} 3600 (fixed -20\%)} \\
\multirow{-2}{*}{\textbf{Unmodelled}} &
  \textbf{RF-DROPO} &
  $(\mu \pm \sigma)$ &
  0.07 $\pm$ 1.01E-02 &
  0.30 $\pm$ 1.70E-03 &
  0.50 $\pm$ 1.10E-03 &
  0.45 $\pm$ 8.00E-04 &
  {\color[HTML]{CB0000} 3600 (fixed -20\%)} \\ \hline
\end{tabular}%
\label{tab:trunk_push_inf}
\vspace{-6pt}
\end{table*}

\subsection{DR for affordability: simplified vs. complex modeling}
\begin{figure}[tb]
    \centering
    \includegraphics[width=1.0\linewidth]{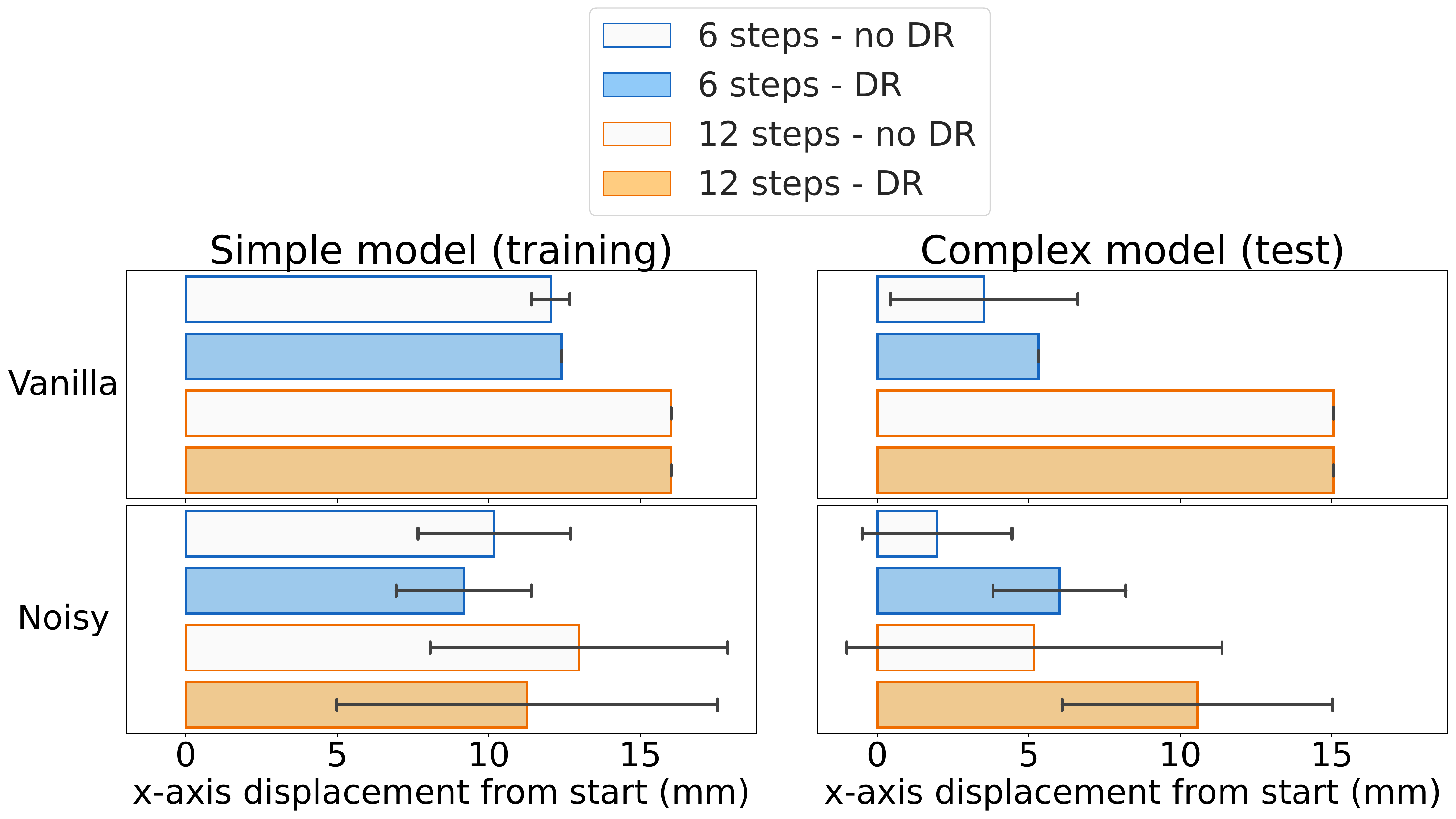}
    \caption{Mean and stdev. displacement achieved by policies trained on the simplified MultiGait model. A mild Gaussian noise with $\sigma=\{0.5mm,100Pa\}$ is optionally applied at test time respectively to the center of mass and the air pressure observations of the MultiGait, compared to the initial state values of $83mm$ and $3500Pa$.}
    \vspace{-0.3cm}
    \label{fig:SimpleToComplex_Fixed_vs_Noisy}
\end{figure}

Due to their modeling complexities, directly training on accurate models of soft robots is not always feasible. This is the case for the MultiGait robot, where a full training procedure would take an estimated time of $\sim 55$days (see Tab.~\ref{tab:multigait_time_complexity}) on 24 parallel CPU threads. For this reason, we investigate the effect of DR to overcome modeling approximations by training with a substantially simpler model and transferring the policy for evaluation to the more accurate MultiGait model.
In particular, we vary the Poisson's Ratio and Young's Modulus of both its constituent materials, and its overall mass at training time. Note how no parameters discrepancy in principle exists between the two models, hence no inference is performed in this analysis. Therefore, domain randomization is applied by means of a random Gaussian noise applied to the nominal values, namely with a fixed $\sigma=\{0.005,0.005,500, 0.01, 10\}$ respectively for the mass, PDMS-PoissonRatio, PDMS-YoungModulus, EcoFlex-PoissonRatio, and EcoFlex-YoungModulus.
We train policies on the simple model only, with both 6-timestep long episodes (i.e. the original version~\cite{schegg2022sofagym}) and 12-timestep long episodes, for 500k timesteps. The former version is heavily limited by the low cardinality of possible strategies\footnote{possible strategies $=6^6=46656$, as only 6 discrete actions are available at each timestep.} and could be simply solved by a brute-force search, failing to motivate the use of RL.
We illustrate the results in Fig.~\ref{fig:SimpleToComplex_Fixed_vs_Noisy}, depicting the final displacement achieved by the robot when rolling out the policies in both the simplified (training) and complex (test) models---3 seeds per training setting.
We notice that policies trained with DR converged to the same strategy as non-DR policies in the training environment (vanilla setting).
More importantly, we observe that DR led to significantly more robust policy behaviors on the complex model, achieving either on par or better average return. The performance gap is exacerbated when testing under noisy observations (5 rollouts per seed), as non-DR policies would sometimes even lead to a backward motion.
Overall, we conclude that Domain Randomization for the MultiGait soft robot design may reduce the training time by 7.9x by leveraging approximated models for training, yet transfer the learned behavior well to more accurate models at evaluation time.

\begin{table}[]
\caption{
TrunkLift: policies trained with a randomized location of the wall learn to exploit environmental constraints and lift up the object's center of mass to a higher elevation. 
}
\resizebox{\columnwidth}{!}{%
\begin{tabular}{cc|ccc}
\hline
\textbf{\begin{tabular}[c]{@{}c@{}}Object\\ Mass {[}$kg${]}\end{tabular}} &
  \textbf{DR Wall} &
  \textbf{\begin{tabular}[c]{@{}c@{}}Environmental\\ Exploitation\end{tabular}} &
  \textbf{Final Object Pose} &
  \textbf{\begin{tabular}[c]{@{}c@{}}Elevation {[}$mm${]} \\ $(\mu \pm \sigma)$\end{tabular}} \\ \hline
\multirow{2}{*}{\textbf{0.1}}  & \textbf{No}  & No  & Unchanged  & 2.34 $\pm$ 4.03  \\ \cline{2-5} 
                               & \textbf{Yes} & Yes & Horizontal & 7.96 $\pm$ 1.67  \\ \hline
\multirow{2}{*}{\textbf{0.05}} & \textbf{No}  & No  & Horizontal & 8.55 $\pm$ 1.27  \\ \cline{2-5} 
                               & \textbf{Yes} & Yes & Vertical   & 12.07 $\pm$ 4.72 \\ \cline{2-5} 
\end{tabular}%
}

\label{tab:env_expl}
\end{table}

\begin{table}[]
\caption{Training and test time complexity on the Multigait simple vs. complex model. Trainings are run on CPU-only, with 24 parallelized environments.}
\vspace{-4pt}
\centering
\resizebox{0.85\columnwidth}{!}{%
\begin{tabular}{l|c|c|}
\cline{2-3}
 &
  \multicolumn{1}{l|}{\textbf{Simple model}} &
  \multicolumn{1}{l|}{\textbf{Complex model}} \\ \hline
\multicolumn{1}{|l|}{\textbf{Training (500k timesteps)}} &
  7 days &
  $\sim$55 days \\ \hline
\multicolumn{1}{|l|}{\textbf{Test (12 timesteps)}} &
  1 min &
  8 minutes \\ \hline
\end{tabular}%
}
\label{tab:multigait_time_complexity}
\vspace{-12pt}
\end{table}

\subsection{DR for effectiveness: environment exploitation}
\begin{figure}[tb]
    \centering
    \includegraphics[width=\linewidth]{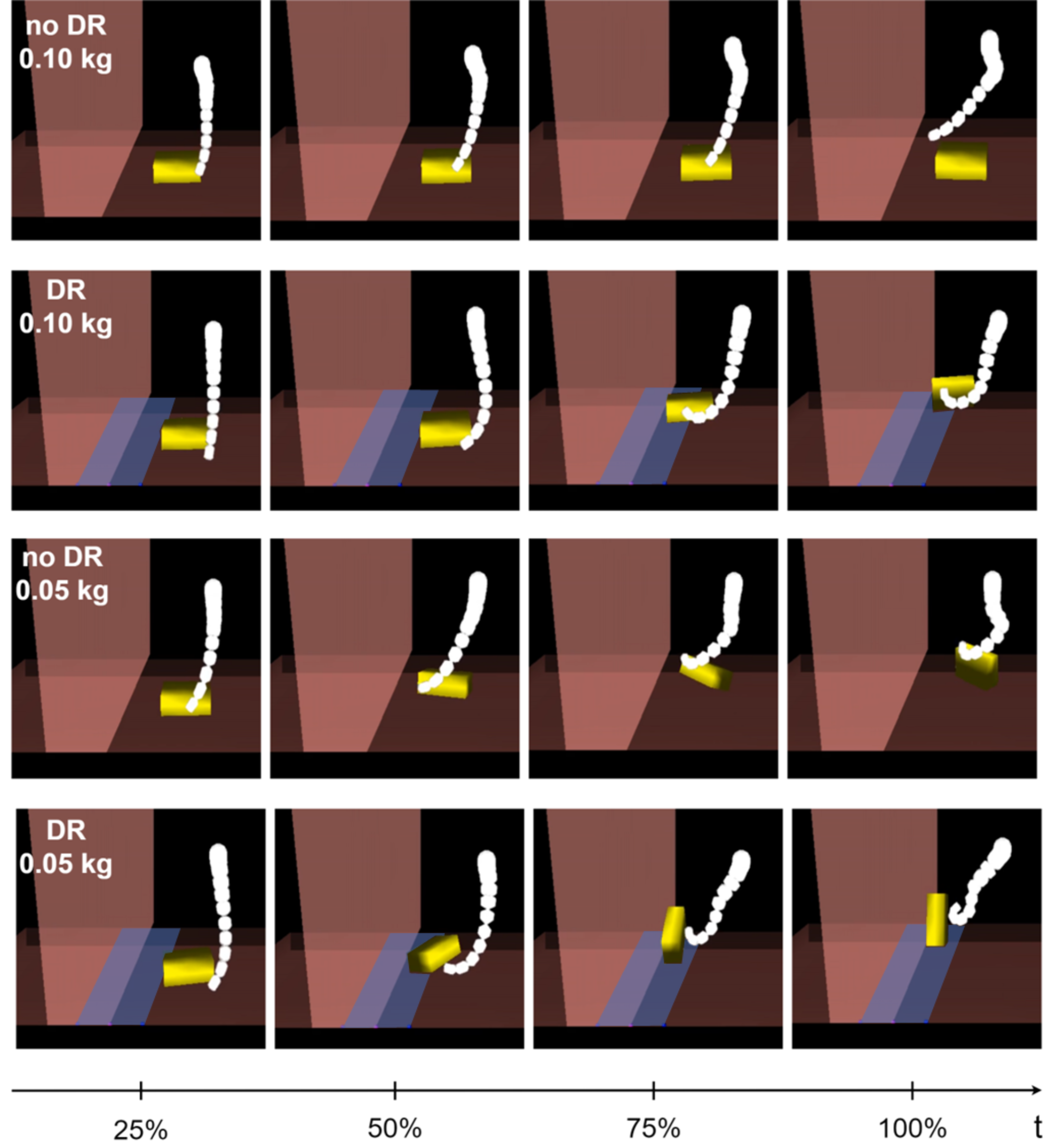}
    \caption{TrunkLift: policies learned with a randomized position of the wall (blue shaded area) learn to exploit it to lift up the object reliably and reach higher average reward.}
    \label{fig:trunkwall_seq_1}
    \vspace{-12pt}
\end{figure}

Motivated by the intrinsic compliance of soft robot bodies, we finally investigate the ability of RL policies to adapt to an unknown environment and explore it in an effective manner. In particular, our analysis revolves around the effects of training on randomized properties of the simulator which, besides allowing robustness to parameter discrepancies and modeling errors, may bias soft agents towards learning more effective strategies.
Indeed, note how soft robots offer a much broader spectrum of potential solutions for completing a task, due to their infinite-DoF nature.
For the sake of this analysis, we consider the TrunkLift setup as a representative example of a task which may be solved through different strategies---e.g. exerting a force on the object's longitudinal side or leverage the nearby wall to lift it up from the shorter side. Arguably, the latter strategy would allow more effective execution under different circumstances---such as different object masses and friction coefficients---and would require inferior effort. One could inject prior knowledge through reward shaping and bias policy search towards specified goals. Instead, we simply randomize the position of the wall at training time, \textit{without} including it among the agent observations, to encourage a more effective exploration of the environment. Note how, in accordance with Sec.~\ref{subsec:dr_background}, this effectively corresponds to an instance of Domain Randomization, forcing the agent to find a \textit{robust} strategy in response to varying wall positions at the start of each episode---which are unknown to the agent. Notably, the uniform randomization range used (see blue shaded area in Fig.~\ref{fig:trunkwall_seq_1}) does not affect the object at its initial location, even in the right-most case.
A tabular description of the experimental analysis is reported in Tab.~\ref{tab:env_expl}, highlighting how the lift-up learned strategy is affected by the introduction of DR. In particular, we observe that for a heavier object mass ($0.1kg$), a feasible solution that does not exploit the wall may not be found, and only policies trained with DR reliably converged to strategies that leveraged the available environmental constraint---$3$ training repetitions.
Note how all policies are tested with the wall location fixed to the nominal location, making it a fair comparison with non-DR policies.
Intuitively, decreasing the mass of the object makes it easier to solve the task, and resulted in both settings (with vs. without DR) finding a way to lift up the object. However, the DR-trained policies converged to the most favorable strategy, as shown in Fig.~\ref{fig:trunkwall_seq_1}.

\section{CONCLUSIONS}
\label{sec:conclusions}
In this work, we provide evidence on how Domain Randomization can improve the robustness, affordability and effectiveness of closed-loop RL policies for soft robot control.
In particular, we propose a novel method to automatically infer posteriors over simulator dynamics parameters given a single trajectory collected on the target domain. We demonstrate that such distributions can be used to train control policies robust to parameter discrepancies and unmodeled phenomena (TrunkReach, TrunkPush).
Additionally, we show that DR with a fixed Gaussian distribution can be used to learn on simplified dynamics models of deformable objects, with a drastic reduction in training time (MultiGait).
Finally, we report evidence that randomizing the agent's surrounding environment may implicitly induce better exploration, and, e.g. lead to exploitation of environmental constraints for optimal performance (TrunkLift).
Future directions of work should thoroughly analyse the deployment of DR-trained policies on real-world setups, closing the sim-to-real loop for soft robot control in contact-rich tasks.


\bibliographystyle{IEEEtran}
\bibliography{root}

\end{document}